\pdfoutput=1
\documentclass[11pt]{article}

\usepackage{ACL2023}

\usepackage{times}
\usepackage{latexsym}

\usepackage[T1]{fontenc}
\usepackage[utf8]{inputenc}

\usepackage{microtype}

\usepackage{graphicx}
\usepackage{multicol}
\usepackage{multirow}
\usepackage{amsmath}
\usepackage{amssymb}
\usepackage{bbding}
\usepackage{subfigure}
\usepackage{colortbl}
\usepackage{pifont}
\usepackage{bbm}
\usepackage{makecell}
\usepackage{bbding}
\usepackage{mathrsfs}
\usepackage{bm}
\usepackage{CJKutf8}
\usepackage{booktabs}
\usepackage{tabu}

\usepackage[linesnumbered,ruled,vlined]{algorithm2e}

\newcommand{\ourmethod}{\textsc{xCoT}}
\newcommand{\xicl}{xICL}
\newcommand{\randomCoT}{Random-CoT}
\newcommand{\msampling}{mSampling}
\newcommand{\xdistill}{xDistill}
\newcommand{\dataset}{\textsc{xCoT-Instruct}}
\title{\ourmethod{}: Cross-lingual Instruction Tuning for Cross-lingual Chain-of-Thought Reasoning}

\author{
  Linzheng Chai\textsuperscript{\rm 1},
  Jian Yang\textsuperscript{\rm 1}\thanks{\ \ Corresponding Author.},
  Tao Sun\textsuperscript{\rm 1}, 
  Hongcheng Guo\textsuperscript{\rm 1}, 
  {\bf Jiaheng Liu}\textsuperscript{\rm 1}, 
  \\{\bf Bing Wang}\textsuperscript{\rm 1}, 
  {\bf Xiannian Liang}\textsuperscript{\rm 1}, 
  {\bf Jiaqi Bai}\textsuperscript{\rm 1}, 
  {\bf Tongliang Li}\textsuperscript{\rm 3}, 
  {\bf Qiyao Peng}\textsuperscript{\rm 2}, 
  {\bf Zhoujun Li}\textsuperscript{\rm 1}\\
  \textsuperscript{\rm 1}State Key Lab of Software Development Environment, Beihang University \\ 
  \textsuperscript{\rm 2}School of New Media and Communication, Tianjin University \\ 
  \textsuperscript{\rm 3}Beijing Information Science and Technology University \\
  \{challenging, jiaya, buaast, hongchengguo, liujiaheng, bingwang, xnliang\}@buaa.edu.cn; \\
  \{bjq, lizj\}@buaa.edu.cm; qypeng@tju.edu.cn; tonyliangli@bistu.edu.cn; \\
 }
\begin{document}
\begin{CJK*}{UTF8}{gbsn}
\maketitle

\begin{abstract}
Chain-of-thought (CoT) has emerged as a powerful technique to elicit reasoning in large language models and improve a variety of downstream tasks. CoT mainly demonstrates excellent performance in English, but its usage in low-resource languages is constrained due to poor language generalization. To bridge the gap among different languages, we propose a cross-lingual instruction fine-tuning framework (\ourmethod{}) to transfer knowledge from high-resource languages to low-resource languages. Specifically, the multilingual instruction training data (\dataset{}) is created to encourage the semantic alignment of multiple languages. We introduce cross-lingual in-context few-shot learning (\xicl{}) to accelerate multilingual agreement in instruction tuning, where some fragments of source languages in examples are randomly substituted by their counterpart translations of target languages. During multilingual instruction tuning, we adopt the randomly online CoT strategy to enhance the multilingual reasoning ability of the large language model by first translating the query to another language and then answering in English. To further facilitate the language transfer, we leverage the high-resource CoT to supervise the training of low-resource languages with cross-lingual distillation. Experimental results on previous benchmarks demonstrate the superior performance of \ourmethod{} in reducing the gap among different languages, highlighting its potential to reduce the cross-lingual gap\footnote{The dataset and code will be released.}.

\end{abstract}

\section{Introduction}
Recent advancements in Large Language Models (LLMs) \cite{llama,llama2,gpt3.5,gpt4,qwen} in natural language processing (NLP) have intensively engaged the interests of researchers. LLMs \cite{cot,multimodal_cot,zeroshot_cot} are further equipped with the chain-of-thought (CoT) technique to gain impressive performance in complex reasoning tasks, where LLMs first produce intermediate reasoning steps and infer the final answer.

%%%%%%%%%%%%%%%%%%%%%%%%%%%%%%%%%%%%%%%%%%%%%%%%%%%%%%%%%%%
\begin{figure}[t]
\centering
\vspace{20pt}
\includegraphics[width=1.0\linewidth]{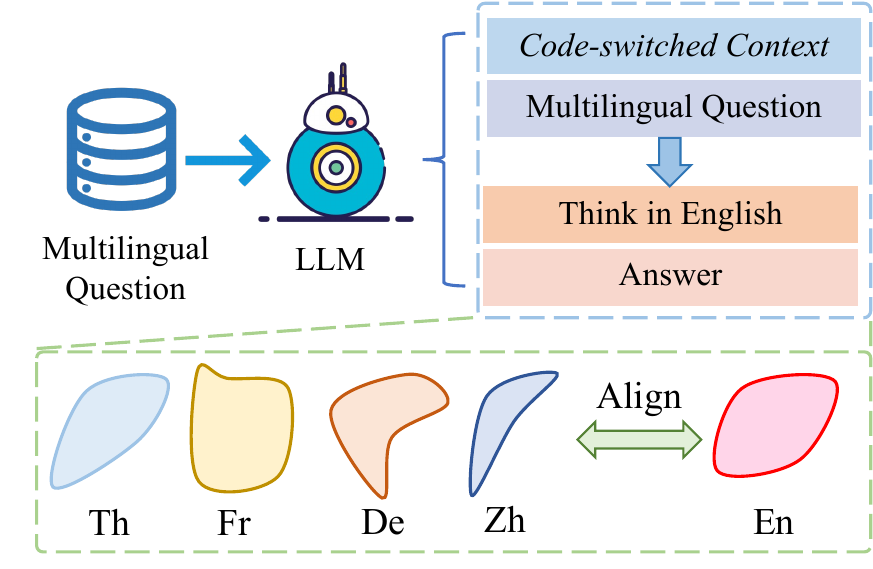}
\caption{Illustration of \ourmethod{}. The cross-lingual instruction tuning is used to align representations of different languages.}
\vspace{-20pt}
\label{intro}
\end{figure}
%%%%%%%%%%%%%%%%%%%%%%%%%%%%%%%%%%%%%%%%%%%%%%%%%%%%%%%%%%%

However, existing studies related to the CoT methods are
mainly constrained in high-resource languages (e.g. English) and deliver little consideration into multilingual scenarios. Recent works \cite{mgsm,crosslingual_prompt,msvamp} endeavor to simply use prompt engineering to improve the language generalization ability of the model without any fine-tuning. These prompt-based methods ignore the potential of representation-based cross-lingual alignment derived from the cross-lingual supervised fine-tuning (cross-lingual SFT). Supervised fine-tuning has been shown to perform at a satisfactory level across various tasks, such as FLAN \cite{flan} and InstructGPT \cite{instruct_gpt}. Therefore, \textit{how to encourage cross-lingual alignment in supervised fine-tuning still requires further exploration.}

%%%%%%%%%%%%%%%%%%%%%%%%%%%%%%%%%%%%%%%%%%%%%%%%%%%%%%%%%%%
\begin{figure*}[htp]
\begin{center}
	\includegraphics[width=0.9\textwidth]{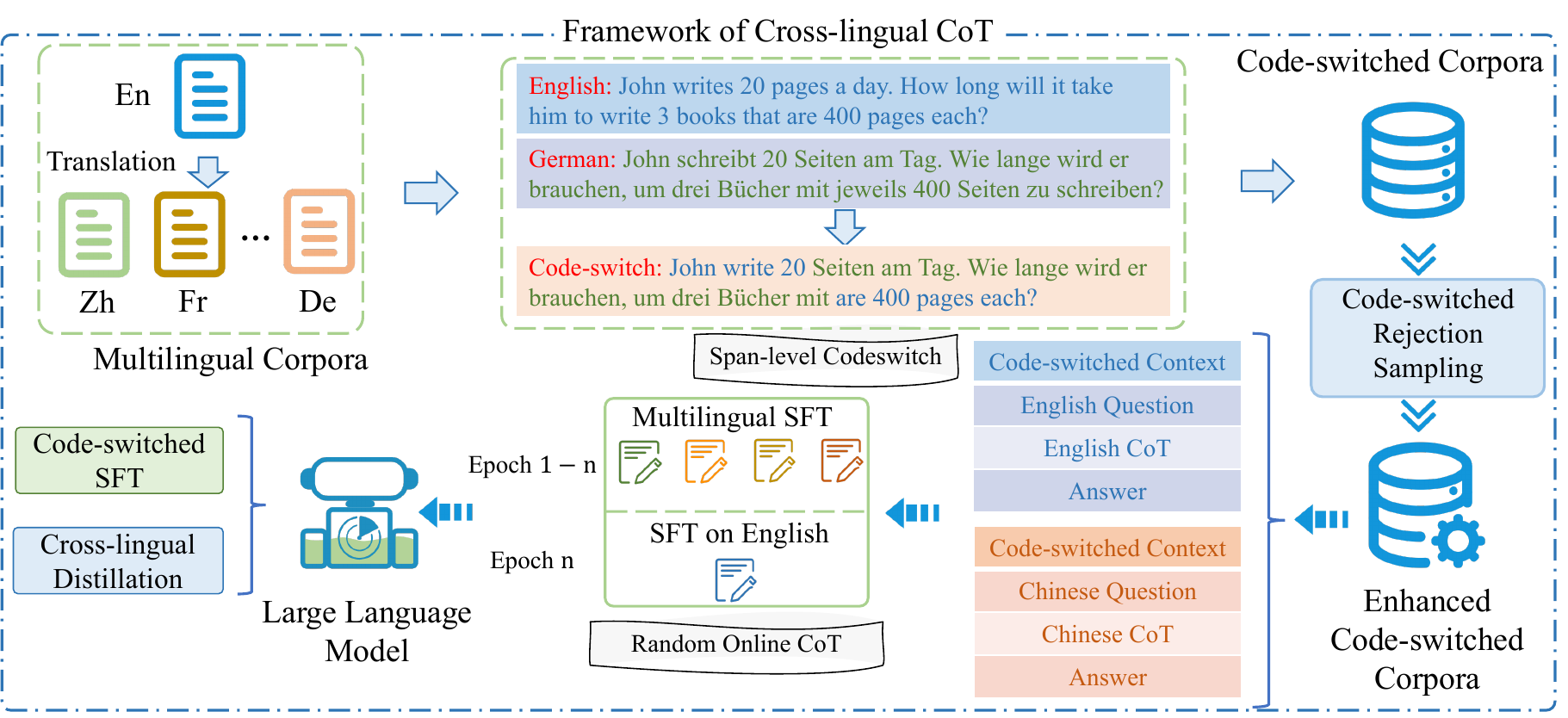}
        \caption{Overview of \ourmethod{}. The cross-lingual in-context few-shot learning (xICL) encourages multilingual alignment in instruction tuning, where the query in the example is mixed with different language tokens. During multilingual instruction tuning, the randomly online CoT strategy (\randomCoT{}) is used to promote the multilingual reasoning ability of LLM and then answer in English. Finally, we leverage the high-resource CoT to supervise the training of low-resource languages with cross-lingual distillation.}
	\label{framework}
	\vspace{-10pt}
\end{center}
\end{figure*}
%%%%%%%%%%%%%%%%%%%%%%%%%%%%%%%%%%%%%%%%%%%%%%%%%%%%%%%%%%%
To minimize the gap among different languages, we propose a \textbf{C}ross-lingual \textbf{C}hain-\textbf{o}f-\textbf{T}hought reasoning (\ourmethod{}) framework using cross-lingual supervised instruction fine-tuning. Specifically, we first construct the multilingual instruction training data (\dataset{}) by translating English to other languages. Then, we randomly substitute some fragments of source languages in examples by their counterpart translations of target languages. To transfer high-resource languages to low-resource languages, we mix the tokens of the source and target language in the same query to enable the LLMs to handle different languages. The code-switched examples and the query can be applied to cross-lingual in-context learning in supervised instruction tuning.
During multilingual instruction tuning, we adopt the randomly online CoT strategy to enhance the multilingual reasoning ability of the large language model by first translating the query to another language and then answering in English. To further facilitate the language transfer, we leverage the high-resource CoT to supervise the training of low-resource languages with cross-lingual distillation. Experimental results on previous benchmarks demonstrate the superior performance of \ourmethod{} in reducing the gap among different languages, highlighting its potential to reduce the cross-lingual gap.

Extensive experiments of \ourmethod{} are evaluated on multilingual benchmarks MGSM of 11 languages and MSVAMP of 10 languages. The results demonstrate that our proposed method consistently achieves state-of-the-art performance across all languages, notably surpassing strong baseline by an average margin of 15\%. The contributions in this work are summarized as follows: (1) We construct the multilingual instruction data to transfer knowledge of high-resource languages into low-resource languages. The training data is further augmented by cross-lingual in-context learning, where a piece of code-switched demonstration context and the current query are concatenated as the input for LLM. (2) During training, we propose the random online CoT (\randomCoT{}), which first randomly translates the query into other languages and then answers in English. (3) To align the representations of different languages, we propose cross-lingual knowledge to align the output distribution given the queries of different languages using Kullback–Leibler divergence.

\section{Cross-lingual CoT Reasoning}
%$e=\{q^{(b)},c^{(b)},a^{(b)}\}_{b=1}^{B}$
Given the query $q=(q_1\dots,q_n)$ of language $L_{i}$, the large language model (LLM) $\mathcal{M}$ outputs the corresponding answer $a=(a_1,\dots,a_n)$ of language $L_{j}$, where $m$ and $n$ are lengths of prompt and answer in a sample $(q, a)$. $L_{i}$ and $L_{j}$ are source and target language, where $L_{all}=\{L_{k}\}_{k=1}^{K}$ and $K$ is number of languages.
LLM further enhances the task performance by chain-of-thought reasoning, where the chain-of-thought examples of sequences $c=(c_1,\dots,c_t)$ are added into the exemplars of prompting. The high-quality rationales $c$ comprised of a series of intermediate natural language reasoning steps provide helpful suggestions for the final output. Given multiple chain-of-thought examples  as demonstrations and the original prompt $q$ of the target language as a whole, the problem definition of cross-lingual CoT is described as:
\begin{MiddleEquation}
\begin{align}
    P(a|q,c) = \prod_{j=1}^{n}P(a_j|a_{<j};q,c,\mathcal{M})
    \label{problem_definition}
\end{align}
\end{MiddleEquation}where $q$ (question) and $c$ (corresponding exemplars) are concatenated as a whole $p$ to predict the answer denoted as $P(a|p)$. Driven by the CoT demonstrations $c$, the LLM first generates the intermediate steps and then outputs the final answer $a$. 
% \begin{MiddleEquation}
% \begin{align}
%     P(a|q) = \prod_{i=1}^{t}P(c_i|p,c_{<i},\mathcal{M})\prod_{j=1}^{n}P(a_j|p,a_{<j},\mathcal{M})
%     \label{problem_definition}
% \end{align}
% \end{MiddleEquation}

\section{\ourmethod{}}
\subsection{Model Overview}
Figure \ref{framework} describes the overall framework of our method \ourmethod{}. Specifically, the cross-lingual in-context few-shot learning (xICL) encourages multilingual alignment in instruction tuning, where the query in the example is mixed with different language tokens. During multilingual instruction tuning, the randomly online CoT strategy (Random-CoT) is used to promote the multilingual reasoning ability of LLM and then answer in English. Finally, we leverage the high-resource CoT to supervise the training of low-resource languages with cross-lingual distillation.
\subsection{\dataset{}}
\paragraph{Data Construction} We create a new multilingual instruction dataset (\dataset{}) for cross-lingual chain-of-thought reasoning, which can be used as the training corpora for multilingual benchmarks, such as MGSM \cite{mgsm} and MSVAMP \cite{msvamp}. We use the multilingual translator to expand the English instruction data\footnote{https://github.com/openai/grade-school-math} into other 10 languages, including German, French, Spanish, Russian, Chinese, Japanese, Thai, Telugu, Bengali, and Swahili. The instruction dataset of each language contains 7.4K samples, where we only translate the query into other languages and retain the response in English to facilitate the cross-lingual transfer. Finally, we obtain the multilingual instruction data $D=\{D^{L_{k}}\}_{k=1}^{K}$ and $(q^{L_i},c^{L_i},a^{L_j}) \in D^{L_i}$, where $D^{L_i}$ is the SFT training data of language $L_{i}$ and the number of the languages is $K$. $D^{L_i}$ contains query $q^{L_i}$ and response $a^{L_j}$ with the corresponding context $c^{L_i}$. $q^{L_i}$ is the query of source language and $a^{L_j}$ is the response of the high-resource language ($L_{j}$ is English in our work). $c^{L_i}=\{q^{L_i}_{b}, a^{L_j}_{b}\}_{b=1}^{B}$ is the context demonstration comprised of $B$ queries of language $L_{i}$ and the responses of $L_{j}$. For each language, we construct about 22K data context demontstration samples.

\begin{algorithm}[t]
\DontPrintSemicolon
\KwIn{
    Multilingual Instruction Dataset: $D$; \\
    Multilingual LLM: $\mathcal{M}$; \\
    Maximum supervised fine-tuning step: $T$; \\
    Batch size: $B$; \\
    Target language set: $\mathcal{L}_{all}=\{L_{k}\}_{k=1}^{K}$;
}
\KwOut{Fine-tuned LLM: $\mathcal{M}$}
$t \gets 0$\;
\While{$t \leq T$;}{
   Random sampled batch $\mathcal{B} \in D$\;
   \For{$k \gets 1$ \textbf{to} $B$;}{
      $(c^{L_{i,j}}, q^{L_{i}}, a^{L_{j}}) \gets \mathcal{B}$\;
      $L_{k} \sim U(\mathcal{L}_{all})$ $(L_{k} \ne L_{j})$\; 
      $q^{L_{k}} \gets \mathcal{M}([c^{L_{i,j}}, q^{L_{i}}, \mathbf{y}_k])$\;
      \tcp*[l]{Translate $q^{L_{i}} \rightarrow q^{L_{k}}$}
      $a^{L_{j}} \gets \mathcal{M}([c^{L_{i,j}}, q^{L_{i}}, q^{L_{k}}, \mathbf{y}_k])$\;
      \tcp*[l]{Answer in language $L_{j}$}
      $\mathcal{B} \gets \mathcal{B} \cup (\mathbf{x}_k^\prime, \mathbf{y}_k, t_k)$\;
    }
    \text{Optimize } $\mathcal{M}$ \text{with} $\mathcal{B}$\;
    $i \gets i + 1$\;
}
\Return{$\mathcal{M}$}\;
\caption{\label{alg_robt}Random Online CoT}
\label{random_cot}
\end{algorithm}
\paragraph{Cross-lingual Instruction Tuning}
Given the cross-lingual instruction corpora $D=\{D^{L_{k}}\}_{k=1}^{K}$, where $D$ contains $K$ languages and $L_{all}=\{L_{k}\}_{k=1}^{K}$. The LLM is jointly trained on the union of the multilingual corpora $D$:
\begin{SmallEquation}
\begin{align}
\begin{split}
    \mathcal{L}_{x} &=-\sum_{i=1}^{K} \mathbb{E}_{c^{L_i},q^{L_{i}},a^{L_{j}} \sim D_{L_{i}}} \left[ \log P(a^{L_{j}}|q^{L_{i}}, c^{L_{i}}; \mathcal{M}) \right] 
    \label{xsft}
\end{split}
\end{align}
\end{SmallEquation}where $q^{L_{i}}$ is the query of the language $L_{i}$ and $a^{L_{j}}$ is the response of language $L_{j}$.

\subsection{Cross-lingual In-context Learning}
To encourage cross-lingual alignment across different languages, we construct the code-switched query by replacing the spans of the source query with the counterparts of the target language.

\paragraph{Code-Switched Sequence}
Given a bilingual query $(q^{L_{i}}, q^{L_{j}})$ with the source language query $q^{L_{i}}=\{q^{L_{i}}_1,\dots,q^{L_{i}}_m\}$ of $m$ tokens and the target translation $q^{L_{j}}=\{y^{L_{j}}_1,\dots,y^{L_{j}}_n\}$ of $n$ tokens, we create the code-switched sequence $q^{L_{i,j}}$ by substituing the phrase $q^{L_{i}}_{j}$ with counterpart translation $q^{L_{j}}_{v_1:v_2}$, where $q^{L_{i}}_{v_1:v_2}$ is the target translation of source piece $q^{L_{j}}_{u_1:u_2}$. $q^{L_{i}}_{u_1:u_2}$ denotes the phrase in $q^{L_{i}}$ from the $u_{1}$-th token to the $u_{2}$-th token and $q^{L_{j}}_{v_{1}:v_{2}}$ denotes the phrase in $q^{L_{j}}$ from the $v_{1}$-th token to the $v_{2}$-th token ($1 \leq v_1 \leq v_2 \leq n$). For each phrase in code-switched in $q^{L_{i,j}}$, it comes from source phrase $q^{L_{i}}_{u_1:u_2}$ or target phrase $q^{L_{j}}_{v_1:v_2}$. The proportion of the source words in the code-switched sequence $q^{L_{i, j}}$ is denoted as $\alpha$. $L_{i,j}$ contains $q^{L_{i/j}}$ (source sentence with target tokesn) and $q^{L_{i/j}}$ (target sentence with target tokesn).

Specifically, the code-switched sequence can be created in two ways:
(1) $q^{L_{i/j}}$ (source sentence with target tokens): most tokens in $q^{L_{i/j}}$ derive from $q^{L_{i}}$, where some source phrases $q^{L_{i}}_{u_1:u_2}$ are substituted by their target counterpart phrases $q^{L_{i_{2}}}_{v_1:v_2}$ ($\alpha \ge 0.5$).
(2) $q^{L_{j/i}}$ (target sentence with source tokens): most tokens in $q^{L_{j,i}}$ derive from $q^{L_{j}}$ where some target phrases $q^{L_{j}}_{v_1,v_2}$ are substituted by their source counterpart phrases $q^{L_{i}}_{u_1,u_2}$ ($\alpha < 0.5$).

\subsection{Random Online CoT}
%https://aclanthology.org/2020.acl-main.148.pdf
To force the model to understand the multilingual queries, we introduce the random online CoT (\randomCoT{}), which first prompts the LLM to translate the query $q^{L_{i_{1}}}$ to another language $q^{L_{i_{2}}}$ and then answer in $a^{L_{j}}$ during the LLM tuning.
\paragraph{Random Online CoT} To scale up to multilingual CoT, we perform online CoT by randomly sampling intermediate languages $L_{i_2}$. Algorithm \ref{random_cot} describes the detail of \randomCoT{}, where given the training instance $(c^{L_{i_{1},j}}, q^{L_{i_{1}}}, a^{L_{j}}) \in D$, we uniformly sample an intermediate language $L_{i_{2}}$ ($L_{i_{2}} \ne L_{i_{1}}$) and prompt LLM first to translate $q^{L_{i_1}}$ to $q^{L_{i_{2}}}$. Although $L_{i_{2}}$ may belong to low-resource languages and the quality of $q^{L_{i_2}}$ may be poor initially, our method still benefits from the translation signal of $q^{L_{i_{1}}} \to q^{L_{i_{2}}}$ by aligning the representations of different languages.
\begin{table*}[t]
    \centering
    \resizebox{0.8\textwidth}{!}{
    \begin{tabu}{llrrrrrrrrrrrr}
    \toprule
    \textbf{Method} & Base Model & En & De & Fr & Es & Ru & Zh & Ja & Th & Te & Bn & Sw & Avg. \\
    \midrule
    \textit{Closed-Source Models}  \\
    \arrayrulecolor{lightgray}\tabucline[0.4pt on 3pt off 3pt]{-} \addlinespace[0.1cm] 
    Native-CoT$^\dag$ & GPT-3.5 & 67.2 & 62.0 & 59.2 & 61.2 & 50.4 & 52.8 & 46.8 & 15.6 & \multicolumn{1}{c}{--} &7.6 &40.0 & 46.3  \\
    Native-CoT$^\dag$ & GPT-4 & \bf 80.0 & \bf 73.6 & \bf 72.0 & \bf 71.2 & \bf 64.0 & \bf 70.0 & \bf 71.6 & \bf 40.4 & \multicolumn{1}{c}{--} & \bf 17.6 & \bf 64.4 & \bf 62.5 \\ 
    \arrayrulecolor{black}\midrule
    \textit{Open-Source Models (7B)} \\
    \arrayrulecolor{lightgray}\tabucline[0.4pt on 3pt off 3pt]{-} \addlinespace[0.1cm] 
    Llama-2$^\dag$~\cite{llama2} & Llama-2 & 43.2 & 37.2 & 34.4 & 32.4 & 28.0 & 22.4 & 15.2 & 4.8 & \multicolumn{1}{c}{--} & 3.2 & 5.2 & 22.6 \\
    RFT$^\dag$~\cite{yuan2023scaling} & Llama-2  & 44.8 & 33.6 & 34.0 & 34.0 & 29.2 & 16.8 & 6.8 & 2.0 & \multicolumn{1}{c}{--} &2.4 & 2.8 & 20.6 \\
    MAmmoTH $^\dag$~\cite{yue2023mammoth}& Llama-2  & 49.6 & 33.2 & 32.8 & 32.4 & 26.0 & 17.2 & 10.8 & 4.8 & \multicolumn{1}{c}{--} & 3.6 & 2.4 & 21.3\\
    WizardMath$^\dag$~\cite{luo2023wizardmath}& Llama-2 & 47.6 & 30.4 & 30.4 & 34.8 & 30.8 &22.4 & 24.0 & 4.0 & \multicolumn{1}{c}{--} & 2.0 & 3.4 & 23.0  \\
    MathOctopus$^\dag$~\cite{chen2023breaking}& Llama-2 & \bf 54.8 & 43.6 & 38.0 & 45.2 & 48.4 & 45.2 & 35.6 & 36.4 & \multicolumn{1}{c}{--} & 33.2 & 38.4 & 41.9 \\
    \midrule
    \ourmethod{}& Bloom &  30.0 &  30.4 & 28.8 & 32.4& 33.6 & 30.0 & 29.6 & 28.4 & 28.4 & 33.2 & 26.8 & 30.1 \\ 
    \ourmethod{}& Llama-2 &  48.4 & \bf 47.2  & \bf 49.6 & \bf 48.8 & \bf 50.0 & \bf 50.0 & \bf 50.0 & \bf 49.2 & \bf 42.8 & \bf 40.4 & \bf 48.4 & \bf 47.7 \\
    \arrayrulecolor{black}\midrule
    \textit{Open-Source Models (13B)}  \\ 
    \arrayrulecolor{lightgray}\tabucline[0.4pt on 3pt off 3pt]{-} \addlinespace[0.1cm] 
    LLama-2$^\dag$~\cite{llama2}& Llama-2 & 50.4 & 42.8 & 40.8 & 45.2 & 39.2 & 32.8 & 25.2 & 6.8 & \multicolumn{1}{c}{--} & 6.0 & 7.6 & 29.7 \\
    RFT$^\dag$~\cite{yuan2023scaling}& Llama-2 & 52.0 & 38.4 &44.8 &46.8 &41.6 &33.6 &26.4&4.4& \multicolumn{1}{c}{--}&3.2 & 3.6& 29.5 \\
    MAmmoth$^\dag$~\cite{yue2023mammoth}& Llama-2 & \bf 56.4 & 45.6 & 39.6 & 50.0 & 36.8 & 31.2 & 19.2 & 5.2 & \multicolumn{1}{c}{--} &3.6& 1.6 & 28.9 \\
    WizardMATH$^\dag$~\cite{luo2023wizardmath}& Llama-2 & 52.8 & 40.4 & 42.0 & 45.6 & 34.4 & 28.0 & 22.0 & 5.6 & \multicolumn{1}{c}{--} & 6.4 & 5.6 & 28.4 \\
    MathOctopus$^\dag$~\cite{chen2023breaking}& Llama-2 & 51.6 & 49.2 & \bf 49.6 & 53.2 & 47.6 & 51.2 & 39.6 & 46.0 & \multicolumn{1}{c}{--} & 42.0 & 46.0 & 47.6 \\
    \midrule
    %v1 \ourmethod{}& Llama-2 &  50.8 & \bf 51.6 & 49.2 & 50.8 & \bf 56.0 & 50.0&\bf 48.8 & \bf 54.0 & \bf 44.8 & \bf 46.8 & \bf 50.0 & \bf 50.2 \\ 
     \ourmethod{}& Llama-2 &  54.4 & \bf 52.4 & 46.4 & \bf 54.8 & \bf 56.8 & \bf 54.0&\bf 49.6 & \bf 50.0 & \bf 47.2 & \bf 50.0 & \bf 51.6 & \bf 51.5 \\ %v2
    \arrayrulecolor{black}\bottomrule
    \end{tabu}}
    \caption{Multilingual evaluation results on the MGSM benchmark. \dag: Results from \cite{chen2023breaking}, for MathOctopus, we uniformly report the performance under xRFT and parallel-training settings.}
    \label{tab:mgsm_main_table}
    \vspace{-10pt}
\end{table*}

\begin{table*}[t]
    \centering
    \resizebox{0.8\textwidth}{!}{
    \begin{tabu}{llrrrrrrrrrrr}
    \toprule
    \textbf{Method} & Base Model & En & De & Fr & Es & Ru & Zh & Ja & Th & Bn & Sw & Avg. \\
    \midrule
    \textit{Closed-Source Models}  \\
    \arrayrulecolor{lightgray}\tabucline[0.4pt on 3pt off 3pt]{-} \addlinespace[0.1cm] 
    Native-CoT$^\dag$ & GPT-3.5 & \bf 81.2 & 73.9 & 78.2 & 74.6 & 70.9 & 78.4 & 74.0 & 46.0 & 14.4 & 68.4 & 66.0 \\
    Native-CoT$^\dag$ & GPT-4 & 80.1 & \bf 78.1 & \bf 83.9 & \bf 81.5 & \bf 77.9 & \bf 78.9 & \bf 74.8 & \bf 68.1 & \bf 31.2 & \bf 75.7 & \bf 73.0 \\ 
    \arrayrulecolor{black}\midrule
    \textit{Open-Source Models (7B)} \\
    \arrayrulecolor{lightgray}\tabucline[0.4pt on 3pt off 3pt]{-} \addlinespace[0.1cm] 
    Llama-2$^\dag$~\cite{llama2} & Llama-2 & 38.8 & 39.0 & 39.1 & 39.2 & 39.1 & 35.2 & 31.6 & 18.2 & 11.5 & 17.2 & 30.9 \\
    RFT$^\dag$~\cite{yuan2023scaling} & Llama-2  & 42.7 & 40.8 & 41.5 & 42.5 & 39.5 & 34.9 & 33.9 & 16.9 & 7.7 & 14.9 & 31.5 \\
    MAmmoTH $^\dag$~\cite{yue2023mammoth}& Llama-2  & 45.1 & 39.6 & 39.9 & 42.9 & 33.7 & 26.8 & 26.7 & 6.3 & 4.3  & 4.2 & 27.0 \\
    WizardMath$^\dag$~\cite{luo2023wizardmath}& Llama-2 & \bf 48.5 & 39.2 & 37.7 & \bf 44.8 & 37.4 & 36.3 & 37.9 & 17.0 &16.1 & 10.3 & 32.5\\
    % MathOctopus$^\dag$~\cite{chen2023breaking}& Llama-2  & \bf 49.9 & 46.5 & 47.3 & 47.6 & 46.6 & 43.3 & 42.7 & 36.2 & 32.9 & 37.7 & 43.1 \\
    MathOctopus$^\dag$~\cite{chen2023breaking}& Llama-2  & 46.8 & 43.1 & \bf 45.3 & 44.5 & 42.1 & 43.2 & 43.2 & \bf 40.5 & 32.8 & \bf 42.3 & 42.4 \\

    % \midrule
    % % \ourmethod{}& Bloom &  30.0 &  30.4 & 28.8 & 32.4& 33.6 & 30.0 & 29.6 & 28.4 & 28.4 & 33.2 & 26.8 & 30.1 \\ 
    % \ourmethod{}& Llama-2 & 47.3 & \bf 44.3 & 42.8 & 44.0 & \bf 43.9 & \bf 43.4 & \bf 43.7 & 39.8 & \bf 38.0 & 41.7 & \bf 42.9  \\
    % \arrayrulecolor{black}\midrule
    % \textit{Open-Source Models (13B)}  \\ 
    % \arrayrulecolor{lightgray}\tabucline[0.4pt on 3pt off 3pt]{-} \addlinespace[0.1cm] 
    % LLama-2$^\dag$~\cite{llama2}& Llama-2 & 50.9 & 46.2 & 47.8 & 46.1 & 47.8 & 43.3 & 41.8 & 23.4 & 13.9 & 19.8 & 38.1 \\
    % RFT$^\dag$~\cite{yuan2023scaling}& Llama-2 & 47.1 & 45.1 & 45.2 & 45.6 & 46.5 & 42.3 & 42.4 & 24.8 & 12.2 & 19.4 & 37.1 \\
    % MAmmoth$^\dag$~\cite{yue2023mammoth}& Llama-2 & 53.4 & 52.3 & 53.8 & 53.9 & 50.7 & 47.7 & 42.2 & 13.7 & 5.0 & 12.9 & 38.6\\
    % WizardMATH$^\dag$~\cite{luo2023wizardmath}& Llama-2 & \bf 56.3 & 48.7 & 49.4 & 50.4 & 43.8 & 37.0 & 29.5 & 16.3 & 13.7 & 12.5 & 35.8 \\
    % MathOctopus$^\dag$~\cite{chen2023breaking}& Llama-2 & 52.9 & 50.5 & 51.5 & 52.8 & 50.2 & 49.2 & 45.8 & 35.7 &  34.1 &  41.9 & 46.5  \\
    % \midrule
    % \ourmethod{}& Llama-2 & 42.7 & 44.2 & 43.7 & 43.7 & 43.9 & 41.2 & 41.1 & 41.2 & 37.5 & 40.6 & 42.0 \\ 
    % \arrayrulecolor{black}\midrule
    % \textit{Open-Source Models (33B)}  \\    
    \arrayrulecolor{black}\bottomrule
    \end{tabu}}
    \caption{Multilingual evaluation results on the MSVAMP benchmark. \dag: Results from \cite{chen2023breaking}, for MathOctopus, we uniformly report the performance under xRFT and parallel-training settings.}
    \label{tab:msvamp_main_table}
    \vspace{-10pt}
\end{table*}

\subsection{Cross-lingual Distillation}
To further augment the cross-lingual instruction tuning, we use the fine-tuned LLM $\mathcal{M}$ to generate the synthetic response of the multilingual queries and then select correct reasoning paths as the augmented dataset $D'$. Finally, our model is trained on the original dataset and augmented dataset $D \cup D'$.

Then, we use the high-resource sample to supervise the low-resource sample to transfer knowledge from high-resource to low-resource languages. Given the parallel high-resource sample $(c^{L_{i,j}}, q^{L_{i}},a^{L_{j}})$ and low-resource sample $(c^{L_{k,j}}, q^{L_{k}},a^{L_{j}})$, the model separately predict the target distribution $p(a^{L_{j}}|c^{L_{i,j}}, q^{L_{i}})$ and $p(a^{L_{j}}|c^{L_{k,j}}, q^{L_{k}})$. Since $q^{L_{i}}$ and $q^{L_{k}}$ are semantically equal, we can leverage distribution $P_{high}=p(a^{L_{j}}|c^{L_{i,j}}, q^{L_{i}})$ to supervise $P_{low}=p(a^{L_{j}}|c^{L_{k,j}}, q^{L_{k}})$ in token level:
\begin{MiddleEquation}
\begin{align}
     \mathcal{L}_{d} = -\frac{1}{n}\sum_{t=1}^{n} \left[P_{high}^{t} \log P_{low}^{t} \right] 
     \label{xdistill}
\end{align}
\end{MiddleEquation}where $P_{high}^t$ and $P_{low}^{t}$ is the $t$-token distribution in answer. Through the token-level cross-lingual distribution, we transfer the high-resource knowledge to low-resource languages.

\section{Experiments}

\subsection{Cross-lingual Supervised Fine-tuning}

For each question in the dataset, we randomly select 2 other questions and corresponding answers as the context. we set $0<\alpha<1$ with a $0.8$ replacement threshold to perform the code-switch operation on the question in context. Specifically, we use English as the source language, and the language corresponding to the question is used as the target language. For the English question, we randomly select an other language as the target language. We implement our model based on Llama-2-7B, Llama-2-13B, and Bloom-7b1. We finetune these models with 3 epochs and use a cosine scheduler with a learning rate of 2e-5 and set 3\% warm up. For cross-lingual distillation, we set $\beta=0.3$.

\subsection{Evaluation}
To comprehensively assess the cross-lingual proficiency of \ourmethod{}, we evaluate the method using the MGSM \cite{mgsm} benchmark, which extends the English GSM8K \cite{gsm8k} dataset into ten typologically varied languages through the manual translation of problems. To conduct a thorough and wide-ranging evaluation of the multilingual mathematical problem-solving skills, we have also created an additional out-of-domain test dataset called MSVAMP \cite{msvamp}, originating from the SVAMP \cite{svamp} dataset. This dataset incorporates mathematical problems in 10 different languages, initially translated using machine translation and subsequently refined through careful human review and correction for accuracy and nuance. Finally, our method is evaluated on MGSM \cite{mgsm} and MSVAMP \cite{msvamp} with the accuracy metric. In the experiments, we report the accuracy of all methods.

\subsection{Baselines}
% The proposed methods are compared with (i) closed-source models such as GPT-3.5-Turbo [47], PaLM [11]; 
% (ii) open-source models such as LLaMA-1 [61], LLaMA-2 [62]; 
% (iii) Supervised Fine-Tuning (SFT), which uses the training set of the original GSM8K or MATH datasets; 
% (iv) Rejection sampling Fine-Tuning (RFT) [69] generates and collects correct reasoning paths as augmented data for fine-tuning; 
% (v) WizardMath [38] which generates samples and trains two reward models using ChatGPT 1 to select samples for fine-tuning.

\ourmethod{} are mainly compared with:
(1) close-source LLM GPT-3.5, GPT-4; (2) open-source models Llama-2 and Bloom. \ourmethod{} primarily conduct experiment base on Llama-2 and compare with other Llama-2 based methods RFT, MathOctopus, MAmmoTH, WizardMath, etc. Futhermore, we select Bloom as base model to explore the performance of \ourmethod{} when combined with multilingual LLM.

\subsection{Main Results}
\paragraph{MGSM} Table \ref{tab:mgsm_main_table} presents the results of our method and previous baselines on MGSM of 11 languages, including En, De, En, Fr, Es, Ru, Zh, Ja, Th, Te, Bn, Sw. Compared to the open-source baseline Llama-2, MAmmoTH \cite{yue2023mammoth} trained with a union of chain-of-thought (CoT) and program-of-thought (PoT) gains strong improvement. Our method significantly outperforms the previous strong baseline MAmmoTH by an average of points. It can prove that our method can leverage cross-lingual in-context learning (\xicl{}) and random online CoT (\randomCoT{}) to encourage alignment across different languages.

\paragraph{MSVAMP} Table \ref{tab:msvamp_main_table} compares the performance of our method with previous relevant methods on MSVAMP of 10 languages. The recent strong multilingual baseline MathOctopus beats the previous baselines MAmmoth and WizardMath with the help of the multilingual instruction dataset MGSM8KInstruct. Further, our proposed method gains the best performance of $42.9$ points in 7B level across all languages, demonstrating that our proposed framework strengthens transferability from the high-resource languages to all other languages.

%%%%%%%%%%%%%%%%%%%%%%%%%%%%%%%%%%%%%%%%%%%%%%%%%%%%%%%%%%%%%%%%%%%%%%%%%%%%%%%%%%%%%%%%%
\begin{table}[!t]
\centering
\resizebox{1.0\columnwidth}{!}{
\begin{tabular}{c|ccccc|ccccc|c}
\toprule
&\multicolumn{5}{c|}{Llama-2-7B} & \multicolumn{5}{c|}{Llama-2-13B} \\
% \textbf{Overall}
$k$ & De & Fr & Es & Ru & Zh & De & Fr & Es & Ru & Zh  & Avg.\\
\midrule
10 & 1.68 & 1.67 & 1.68 & 1.80 & 1.69 & 1.98 & 1.96 & 1.98 & 1.98 & 1.96 & 7.22\\
20 & 2.58 & 2.56 & 2.59 & 2.66 & 2.59 & 2.95 & 2.95 & 2.97 & 2.97 & 2.94 & 11.02\\
30 & 3.21 & 3.24 & 3.21 & 3.35 & 3.22 & 3.70 & 3.71 & 3.70 & 3.73 & 3.69 & 13.93\\
50 & 4.32 & 4.29 & 4.34 & 4.47 & 4.35 & 4.93 &4.93  & 4.91 & 4.72 & 4.91 & 18.43\\
% \midrule
% 500 & \multicolumn{10}{c}{18.43} \\
\bottomrule
\end{tabular}}
\caption{Distinct reasoning paths of each language with different sampling times. Different reasoning paths per question are generated by different SFT models with different $k$.}
\label{tab:sampling_path}
\vspace{-10pt}
\end{table}
%%%%%%%%%%%%%%%%%%%%%%%%%%%%%%%%%%%%%%%%%%%%%%%%%%%%%%%%%%%%%%%%%%%%%%%%%%%%%%%%%%%%%%%%%%%

\section{Analysis}
\label{analysis}

\paragraph{Ablation Study}
To verify the effectiveness of each module in our method, we conduct an ablation study by adding modules gradually. The multilingual LLM Llama-7B is first trained on the multilingual corpora \dataset{}, where the model is denoted as {{\large{\ding{172}}}}. Compared to the initial model {\large{\ding{172}}}, the model {\large{\ding{173}}} with the code-switched context in multilingual tuning gains the improvement of $+4.7$ points on average, which shows the usage of \xicl{} in encouraging alignment across different languages. Then, the model {\large{\ding{174}}} is further enhanced with \msampling{} by a large margin $+5.8$ points, where the model generates the multilingual responses and chooses correct reasoning paths as the augmented dataset. During multilingual tuning, our method adopts \randomCoT{} to first translate the query to another language and then answer in English. For the output distribution, the high-resource distribution is used to supervise the low-resource distribution (\xdistill{}).
Putting them all together, we obtain the final model \textbf{\ourmethod{}} ({\large{\ding{176}}}) with $47.7$ points.
Table~\ref{module_ablation} summarizes the results of the ablation study of cross-lingual transfer in different parts, which emphasizes the effectiveness of cross-lingual transfer that can gradually improve performance in different aspects.
%%%%%%%%%%%%%%%%%%%%%%%%%%%%%%%%%%%%%%%%%%%%%%%%%%%%%%%%%%%%%%%%%%%%%%%%%%%%
\begin{table*}[t]
\centering
\resizebox{0.95\textwidth}{!}{
\begin{tabular}{ll|l|cccccccccccc}
\toprule
 ID & Method & En & De & Fr & Es & Ru & Zh & Ja & Th & Te & Bn & Sw & AVG \\
\midrule
 {\large{\ding{172}}} & Llama-2-7B   & 38.4 & 37.2 & 37.6 & 39.2 & 37.6 & 30.4 &  29.6 & 26.8 & 13.2 & 23.2 & 18.4 & 30.1  \\
 {\large{\ding{173}}} & \quad + \textit{\xicl{}}  & 45.6 & 38.4 & 33.6 & 36.4 & 40.4& 37.2 & 32.0 &  32.0 & 26.0 & 30.0 & 32.0 & 34.8    \\
 {\large{\ding{174}}} & \quad + \textit{\msampling{}} & \bf 50.8 & 41.2 & 42.0 & 44.4 & 42.0& 40.4 & 38.0 &  40.8 & 35.6 & 34.8 & 37.6 & 40.6            \\
 {\large{\ding{175}}} & \quad + \textit{\randomCoT{}} & 48.0 & \bf 50.4 & 46.0 & 48.4 & 46.8 & \bf 51.2 &  46.4 & 47.2 & 41.6 &  \bf  41.2 & 46.8 & 46.7      \\ 
 {\large{\ding{176}}} & \quad + \textit{\xdistill{}}  & 48.4 & 47.2  & \bf 49.6 & \bf 48.8 & \bf 50.0 &  50.0 & \bf 50.0 & \bf 49.2 & \bf 42.8 & 40.4 & \bf 48.4 & \bf 47.7    \\
\bottomrule
\end{tabular}
}
\caption{Ablation study on MGSM based on Llama-2-7b. \ourmethod{} is the final model of our method.}
\label{module_ablation}
\vspace{-5pt}
\end{table*}
\begin{table*}[t]
\centering
\resizebox{0.85\textwidth}{!}{
\begin{tabular}{c|c|cccccccccccc}
\toprule
\randomCoT{} Direction  & En &De & Fr & Es & Ru & Zh & Ja & Th & Te & Bn & Sw & Avg. \\
\midrule
$L_{all} \to$ En    & 50.8 & 41.2 &  42.0 &  44.4 &  42.0 & 40.4 &  38.0 &  40.8 & 35.6 & 34.8 &  37.6 &  40.6       \\
\makecell[c]{$L_{low} \to$ En $\cup$ $L_{high} \to L_{high}$} & 50.8 & 45.6 & 49.6 & 46.4 & 49.6 &  41.2 & 42.4 & 42.8 & 37.6 & 38.8 & 42.8 & 44.3  \\
$L_{all} \to L_{all}$ & 46.0 & 46.4 & 45.2 & 50.0 & 48.0& 50.4 & 47.6 & 43.2 & 40.8 & 40.4 & 45.2 & 45.7     \\
$L_{all} \to L_{high}$  & 48.4 & 47.2  & \bf 49.6 & \bf 48.8 & \bf 50.0 &  50.0 & \bf 50.0 & \bf 49.2 & \bf 42.8 & 40.4 & \bf 48.4 & \bf 47.7   \\
\bottomrule
\end{tabular}}
\caption{Translation direction of \randomCoT{}.}
\label{tab:random_cot_analysis}
\end{table*}
%%%%%%%%%%%%%%%%%%%%%%%%%%%%%%%%%%%%%%%%%%%%%%%%%%%%%%%%%%%%%%%%%%%%%%%%%%%%

\paragraph{Cross-lingual Prompting}
%%%%%%%%%%%%%%%%%%%%%%%%%%%%%%%%%%%%%%%%%%%%%%%%%%%%%%%
\begin{table*}[htp]
    \centering
    \resizebox{0.85\textwidth}{!}{
    \begin{tabular}{llrrrrrrrrrrrr}
    \toprule
    \textbf{Base Model}& CoT & En & De & Fr & Es & Ru & Zh & Ja & Th & Te & Bn & Sw & AVG \\ \midrule
    \multirow{3}{*}{Bloom-7B} &En-Context & \bf 32.8 & \bf 31.2 & \bf 30.4 & 30.0 & \bf 34.0 &29.6 & 28.0 & 27.6 & 27.2 &30.4 & 28.4 &29.9  \\
    &Native-Context & \bf 32.8 & 30.0 & 25.6 & 32.0 &31.2 & \bf 30.0 &25.2 &\bf 28.8 & 25.6 &30.0 & \bf 30.4 & 29.2  \\ 
    &Codeswitch-Context & 30.0 & 30.4 & 28.8 & \bf 32.4& 33.6 & \bf 30.0 & \bf 29.6 & 28.4 & \bf 28.4 & \bf 33.2 & 26.8 & \bf 30.1 \\ 
    \arrayrulecolor{lightgray}\midrule
    \multirow{3}{*}{Llama-2-7B} &En-Context & \bf 50.0 & 46.4 & \bf 50.0 & 47.2 & 47.6 & \bf 50.4 & 47.6 & 44.8 & 42.0 & 39.2 & \bf 48.4 & 46.7 \\
    &Native-Context & \bf 50.0 & \bf 48.4 & 44.0 & \bf 48.8 & 47.6 & 46.4 & 45.6 & 45.6 & 42.0 & \bf 41.2 & 44.4 & 45.8  \\ 
    &Codeswitch-Context &   48.4 & 47.2  & 49.6 & \bf 48.8 & \bf 50.0 & 50.0 & \bf 50.0 & \bf 49.2 & \bf 42.8 & 40.4 & \bf 48.4 & \bf 47.7 \\
    \arrayrulecolor{lightgray}\midrule
    \multirow{3}{*}{Llama-2-13B}
    & En-Context & \bf 50.8 & 50.4 & 48.0 & 50.4 & 55.6 & 48.4 & 48.0 & \bf 54.0 & 44.8 & 45.6 & 49.2 & 49.5 \\ 
    
    &Native-Context & \bf 50.8 & \bf 52.0 & \bf 49.2 & 49.2 & \bf 56.0 & \bf 50.0 & \bf 49.6 & 52.8 &  \bf 46.0 & 45.2 & 48.8 & 49.9 \\ 
    
    &Codeswitch-Context & \bf 50.8 & 51.6 & \bf 49.2 & \bf 50.8 & \bf 56.0 & \bf 50.0 & 48.8 & \bf 54.0 & 44.8 & \bf 46.8 & \bf 50.0 & \bf 50.2 \\ 
    \arrayrulecolor{black}\bottomrule
    \end{tabular}}
    \caption{Cross Lingual Promping}
    \label{tab:cross_prompting}
\end{table*}
%%%%%%%%%%%%%%%%%%%%%%%%%%%%%%%%%%%%%%%%%%%%%%%%%%%%%%%%%%%
To trigger the cross-lingual potential capability of LLM, we introduce \xicl{} to force the model to understand the multilingual queries and align their representations. To advance multilingual agreement in instruction tuning, we randomly replace some fragments of source languages in examples with their counterpart translations of target languages for cross-lingual in-context few-shot learning (\xicl{}). Table \ref{tab:cross_prompting} shows the results of \ourmethod{} with English context, native context, and code-switched context on different backbones. The query mixed with different language tokens brings significant improvement in different languages.

\begin{table*}[!h]
\centering
\resizebox{1.0\textwidth}{!}{
\begin{tabular}{cl}
\toprule
\toprule
\bf Spanish Question & \makecell[l]{Dan planta 3 rosales. Cada rosal tiene 25 rosas. Cada rosa tiene 8 espinas. ¿Cuántas espinas hay en total?}\\
\arrayrulecolor{gray}\midrule
Llama-2 & \makecell[l]{There are 3 x 25 = 75 roses.There are 75 x 8 = 600 thorns.\\There are 600 x 2 = 1200 forks. The answer is 1200.\color{red}\ding{56}}\\
\arrayrulecolor{gray}\midrule
\ourmethod{} & \makecell[l]{Question:Dan plants 3 rose bushes. Each rose bush has 25 roses. Each rose has 8 thorns. \\ How many thorns are there in total? Let's think step by step in English. \\There are 25*3=<<25*3=75>>75 roses So there are 75*8=<<75*8=600>>600 thorns. The answer is 600. \color{green}\ding{52}} \\
\arrayrulecolor{black}\midrule
\bf Chinese Question & \makecell[l]{格雷琴有 110 个硬币。金币比银币多 30 个。格雷琴有多少个金币？}\\
\arrayrulecolor{gray}\midrule
Llama-2 & \makecell[l]{Let x be the number of silver coins Gretchen has. Gold coins are 30 more than silver coins, so each gold coin \\ is worth x + 30. The total value of the gold coins is the sum of the value of each gold coin, which is x + 30, \\so the total value is x + 30 * x + 30 = 110. So the equation is 30x + 30 = 110.\\So the value of x is 110 / 30 = 3.5. So Gretchen has 3.5 gold coins. The answer is 3.5.\color{red}\ding{56}}\\
\arrayrulecolor{gray}\midrule
\ourmethod{} & \makecell[l]{Question:Gretchen has 110 coins. She has 30 more gold coins than silver coins. How many gold coins \\ does Gretchen have? Let's think step by step in English. Let G be the number of gold coins Gretchen has. \\ The number of silver coins is G - 30. The total number of coins is G + (G - 30) = 110. \\ Combining like terms, we get G*2 - 30 = 110. Adding 30 to both sides, we get G*2 = 140.\\Dividing both sides by 2, we get G = 70. The answer is 70. {\color{green}\ding{52}}} \\
\midrule
\bf German Question & \makecell[l]{Jimmy hat 2 \$ mehr als doppelt so viel Geld wie Ethel. Wenn Ethal 8 \$ hat, wie viel Geld hat dann Jimmy im Moment?}\\
\arrayrulecolor{gray}\midrule
Llama-2 & \makecell[l]{Jimmy's money is 2 * \$8 = \$16. The answer is 16.{\color{red}\ding{56}}}\\
\arrayrulecolor{gray}\midrule
\ourmethod{} & \makecell[l]{Question:Jimmy has \$2 more than twice the money Ethel has. If Ethel has \$8, how much money does Jimmy have now?\\Let's think step by step in English. Twice the money Ethel has is 2 * 8 = <<2*8=16>>16 dollars. Jimmy has 2 \\ more than twice the money Ethel has, which means he has 16 + 2 = <<16+2=18>>18 dollars. The answer is 18.{\color{green}\ding{52}}} \\
\arrayrulecolor{black}
\bottomrule
\bottomrule
\end{tabular}}
\caption{Comparison between Llama-2 and our proposed method within 7B-level on MGSM. } 
\label{tab:example_study}
\end{table*}
%%%%%%%%%%%%%%%%%%%%%%%%%%%%%%%%%%%%%%%%%%%%%%%%%%%%
\paragraph{Example Study}
Given the queries of different languages, our method prompts LLM to first consider the multilingual query in English and then answer in English. Table \ref{tab:example_study} shows examples of the Spanish, Chinese, and German baseline. We observe that Llama-2 tends to generate incorrect answers for non-English queries. For the German Question ``Jimmy hat 2 \$ mehr als doppelt so viel Geld wie Ethel. Wenn Ethal 8 \$ hat, wie viel Geld hat dann Jimmy im Moment?'', our method first thinks the non-English query in English ``Question:Jimmy has \$2 more than twice the money Ethel has. If Ethel has \$8, how much money does Jimmy have now?''. and then answer in English. It proves that our method can align both query and response across different languages.
\paragraph{Cross-lingual Reasoning Path}
Our multilingual instruction data is augmented by multilingual sampling, where the fine-tuned LLM generates the response and selects the correct path. Table \ref{tab:sampling_path} shows that different languages have a similar number of reasoning paths, which proves that using the cross-lingual CoT successfully transfers reasoning patterns from one language to another language. \ourmethod{} can accumulate all reasoning paths to improve the model performance.

\begin{figure}[t]
    \centering
    \subfigure[Baseline]{
    \includegraphics[width=0.45\columnwidth]{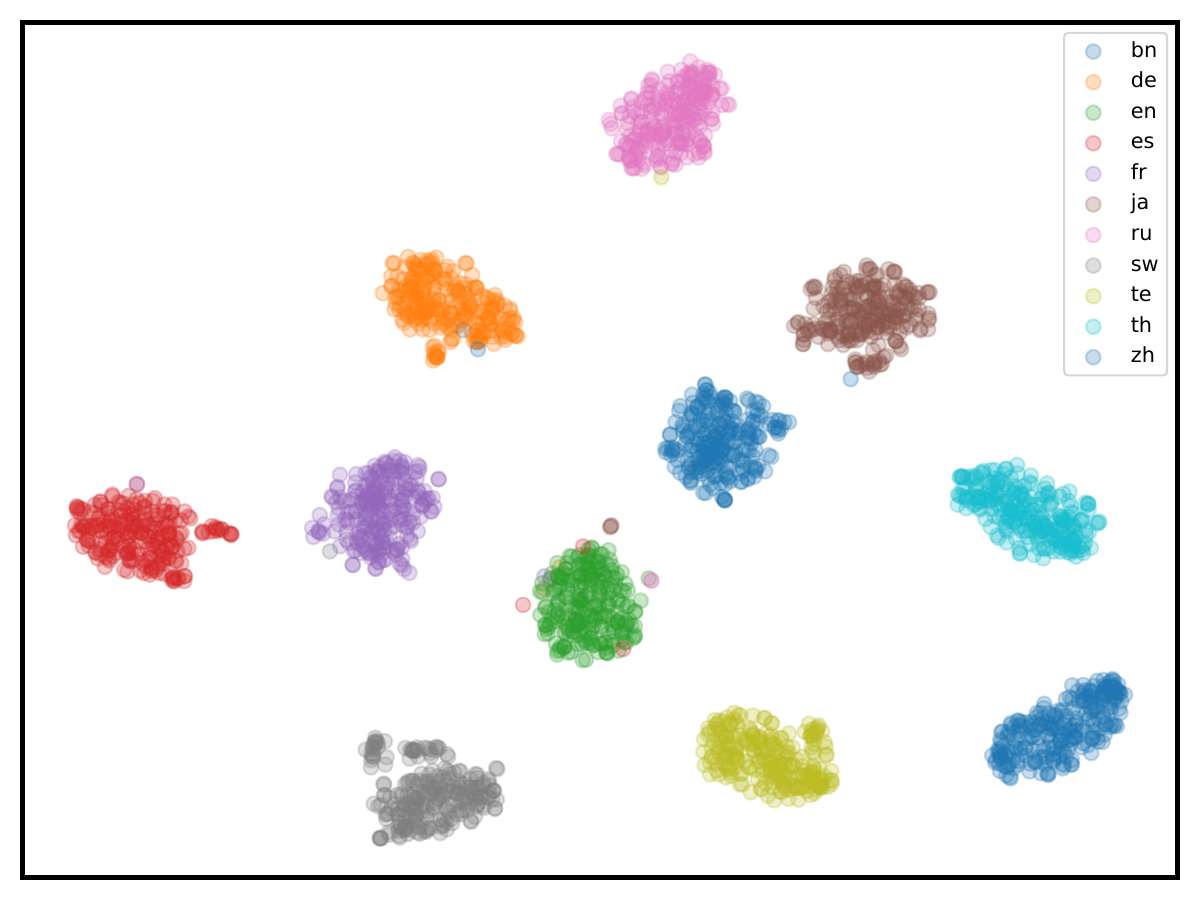}
    \label{tsne_1}
    }
    \subfigure[\ourmethod{}]{
    \includegraphics[width=0.45\columnwidth]{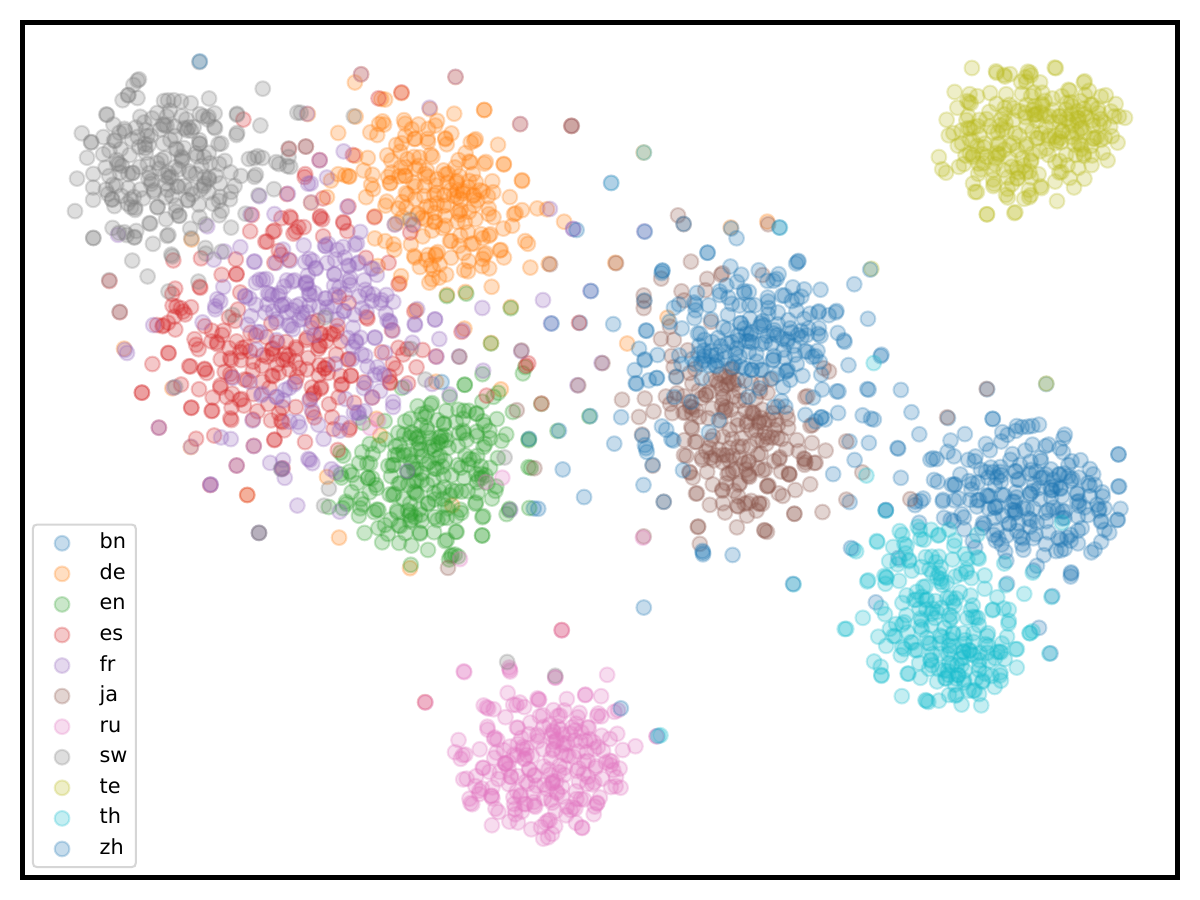}
    \label{tsne_2}
    }
    \caption{(a) and (b) are representations of Llama-7B and our method from the last decoder layer. Each color denotes one language (11 languages in MGSM).} 
    \vspace{-10pt}
    \label{tsne_figures}
\end{figure}

\paragraph{Multilingual Representations}
We randomly select 250 parallel queries with their 2-shot examples of each language in \dataset{} and visualize their representations \cite{t_SNE} of the last Llama decoder layers in Figure \ref{tsne_figures} using our multilingual model fine-tuned on \dataset{} and the multilingual baseline. The first hidden state of the encoder is adopted as the sentence representation. Compared to Figure \ref{tsne_1} of the baseline, different languages become closer and more likely to overlap with each other in Figure \ref{tsne_2} of our method, demonstrating that our method effectively aligns representations of different languages to the shared space.
%%%%%%%%%%%%%%%%%%%%%%%%%%%%%%%%%%%%%%%%%%%%%%%%%%%%%%%%%%%%
%%%%%%%%%%%%%%%%%%%%%%%%%%%%%%%%%%%%%%%%%%%%%%%%%%%%%%%%%%%%

%%%%%%%%%%%%%%%%%%%%%%%%%%%%%%%%%%%%%%%%%%%%%%%
\begin{figure}[h]
\begin{center}
\includegraphics[width=0.8\columnwidth]{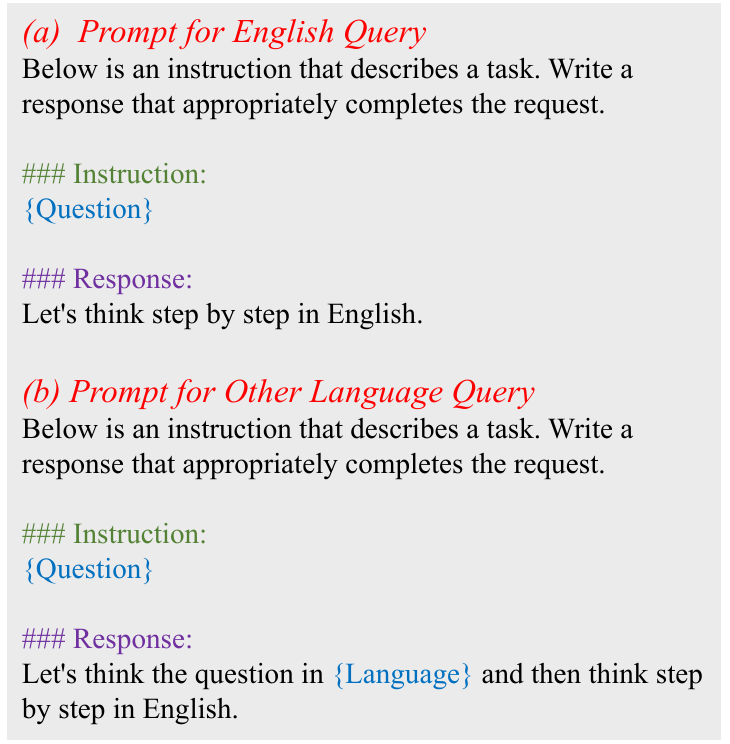}
\caption{The prompt of thinking in English.}
\label{prompt}
\vspace{-15pt}
\end{center}
\end{figure}
%%%%%%%%%%%%%%%%%%%%%%%%%%%%%%%%%%%%%%%%%%%%%%%
\paragraph{Understanding and Reasoning in English}
After the cross-lingual SFT with \randomCoT{}, \ourmethod{} chooses the high-resource language (English) as the auxiliary language to understand and answer the non-English question. In Figure \ref{prompt}, our method uses ``Let's think in step by step in \textit{English}'' to answer the English question. For the non-English question, we adopt ``Let's think the question in \textit{\{Language\}} and then think step by step in \textit{English}'', where \textit{\{Language\}} can be high-resource languages in SFT tuning but we set textit{\{Language\}} as English during inference stage.
To effectively transfer knowledge from high-resource to low-resource languages, we force LLM to understand the query in English and then think in English. 

\paragraph{Analysis in \randomCoT{}}
To facilitate the alignment among different languages, the question of language $L_{i_1}$ is first translated into another language $L_{i_2}$ in SFT tuning. Given the query of language $L_{i_1}$, we can translate another language $L_{i_2}$ ($L_{i_1} \neq L_{i_2}$). The strategy ``$L_{all} \to L_{e}$'' denotes that the $L_{i_1} \in L_{all} \land L_{i_2} = L_{e}$.
Table \ref{tab:random_cot_analysis} shows the results of our method with different \randomCoT{} strategies and the strategy ``$L_{all} \to L_{high}$'' gets the best performance, which can be attributed the language transfer from high-resource to low-resource languages.

\paragraph{Low-resource Setting}
%%%%%%%%%%%%%%%%%%%%%%%%%%%%%%%%%%%%%%%%%%%%%%%%%%%%%%%%%%%
\begin{figure}[t]
\centering
\includegraphics[width=0.75\linewidth]{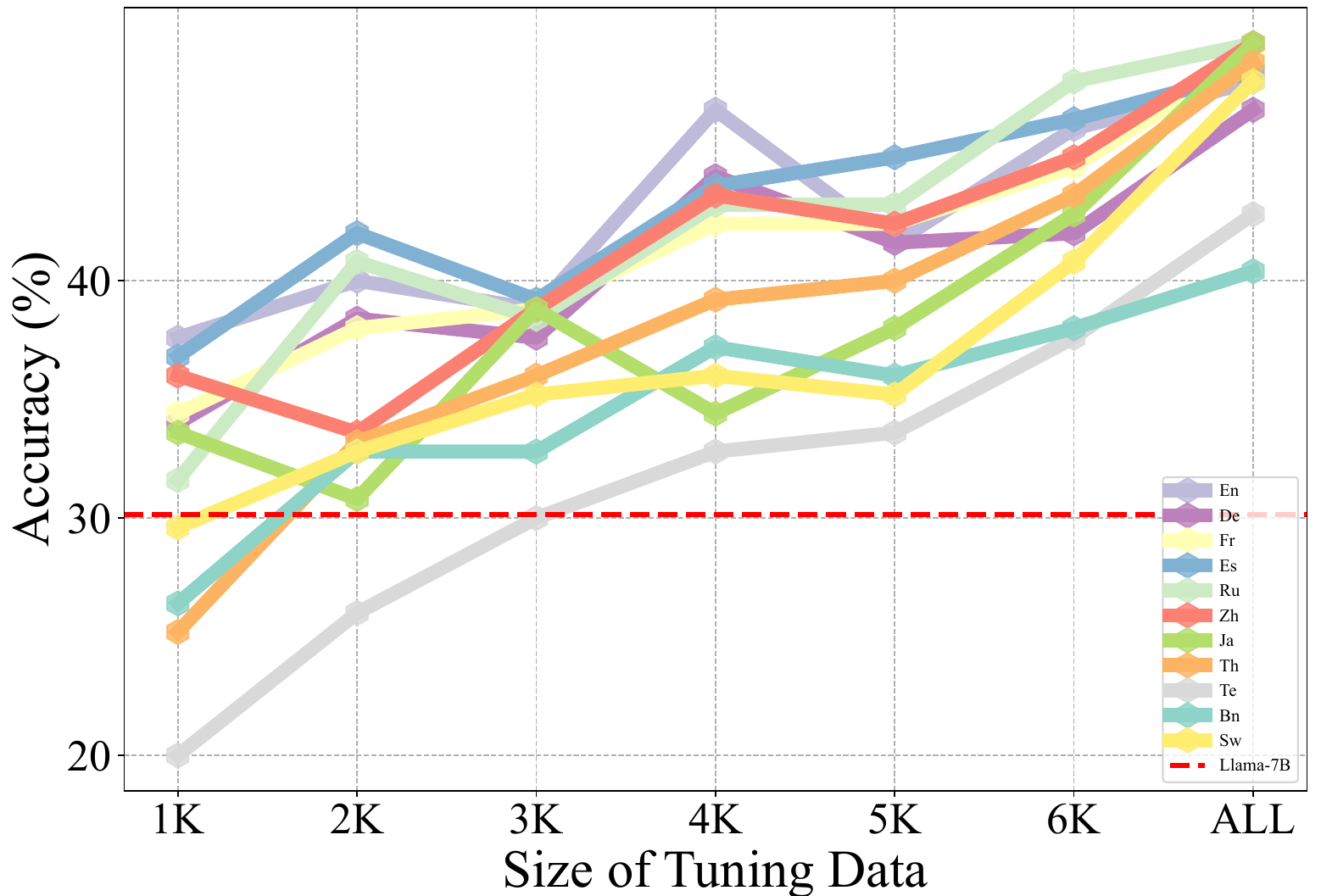}
\caption{Multilingual evaluation results on MGSM with different SFT data size.}
\vspace{-20pt}
\label{low_resource_setting}
\end{figure}
%%%%%%%%%%%%%%%%%%%%%%%%%%%%%%%%%%%%%%%%%%%%%%%%%%%%%%%%%%%
Figure \ref{low_resource_setting} plots the multilingual evaluation results of \ourmethod{} with different SFT data sizes. We observe that our method with nearly 20\% SFT data can still beat the strong baseline Llama-7B, which can attributed to the mutual reinforcement of multiple languages.

\section{Related Work}

\paragraph{Large Language Models} 
Large language models (LLMs) has shown great power in numerous NLP tasks, and as the scale of the model gets larger, LLMs emerge with surprising capabilities~\cite{touvron2023llama,wei2022emergent,glm,owl}, such as following human instructions, in-contextual learning, and reasoning complex tasks. \citet{wei2022chain} found that LLM can solve complex problems efficiently by chain-of-thought prompting strategy (providing some exemplars containing reasoning steps to guide the model to generate intermediate reasoning steps). Moreover, \citet{kojima2022large} found that LLMs can solve complex problems by CoT even without providing exemplars. However, the CoT capability usually requires the model to have a particularly large number of parameters and require massive computational resources. There is also some works~\cite{ho2022large, zhu-etal-2023-solving} that explore the smaller LLMs' CoT capability. In this paper, we focus on the CoT capability for smaller LLMs and further migrate it to multilingual reasoning.

\paragraph{Cross-lingual Transfer} 
Cross-lingual transfer pertains to utilizing labeled data from a resource language to address the challenge of insufficient labeled data in the target language.
Previous works~\cite{xlm,xlmr,alm,xlmt,yang-etal-2023-ganlm} demonstrate that pre-trained models trained on multi-lingual data proficiently perform cross-lingual transfer tasks. These multi-lingual pre-trained models have found extensive application across various downstream NLP tasks, such as multi-lingual translation~\cite{cluster_mnmt,hlt_mt,multi_clinic_trans,um4}, cross-lingual summarization~\cite{cross_sum,under_cross}, cross-lingual information extraction~\cite{con_ner,crop,wu2020Unitrans}. Many LLMs are trained on multilingual data, endowing them with strong cross-linguistic abilities~\cite{scao2022bloom, muennighoff2022crosslingual}. However, the cross-language capability in smaller LLM is not significant, so we augmenting the multilingual reasoning potential of LLMs by employing pseudo training data derived from labeled source-language datasets.

\section{Conclusion}
In this work, we propose a cross-lingual instruction fine-tuning framework (\ourmethod{}) to address the disparity by encouraging alignment among different languages. A cross-lingual instruction dataset (\dataset{}) is first created to align semantically the reasoning capability across various languages. Then, our method incorporates cross-lingual in-context Learning (\xicl{}) to trigger the cross-lingual alignment. During instruction tuning, we adopt random online CoT (\randomCoT{}), which prompts LLM to translate the query into different languages and subsequently provide an English response. To further promote language transfer, we leverage a high-resource CoT to guide low-resource CoT training with cross-lingual distillation (\xdistill{}). Our comprehensive evaluation of established benchmarks showcases the effectiveness of \ourmethod{} in narrowing the multilingual linguistic gap. The results highlight its potential as a robust solution for reducing the cross-lingual divide, setting a new precedent for multilingual language model performance.

% \section*{Acknowledgments}
% This work was supported in part by the National Natural Science Foundation of China (Grant Nos. 62276017, U1636211, 61672081), the 2022 Tencent Big Travel Rhino-Bird Special Research Program, and the Fund of the State Key Laboratory of Software Development Environment (Grant No. SKLSDE-2021ZX-18). 

\bibliography{custom}
\bibliographystyle{acl_natbib}

\clearpage
\end{CJK*}
\end{document}